\documentclass[runningheads]{llncs}
\usepackage{amsmath}
\usepackage{graphicx}
\usepackage{subfigure}
\usepackage{mathrsfs}
\begin{document}
\title{CartoonRenderer: An Instance-based Multi-Style Cartoon Image Translator}
%
%\titlerunning{Abbreviated paper title}
% If the paper title is too long for the running head, you can set
% an abbreviated paper title here
%
\author{Yugang Chen \and
Muchun Chen \and
Chaoyue Song \and Bingbing Ni}
\authorrunning{Y.Chen et al.}
% First names are abbreviated in the running head.
% If there are more than two authors, 'et al.' is used.
%
\institute{ Shanghai Jiao Tong University \\
\email{\{cygashjd,Brunestod,beyondsong,nibingbing\}@sjtu.edu.cn}}
\maketitle              % typeset the header of the contribution
\begin{abstract}
Instance based photo cartoonization is one of the challenging image stylization tasks which aim at transforming realistic photos into cartoon style images while preserving the semantic contents of the photos. State-of-the-art Deep Neural Networks (DNNs) methods still fail to produce satisfactory results with input photos in the wild, especially for photos which have high contrast and full of rich textures. This is due to that: cartoon style images tend to have smooth color regions and emphasized edges which are contradict to realistic photos which require clear semantic contents, i.e., textures, shapes etc. Previous methods have difficulty in satisfying  cartoon style textures and preserving semantic
contents at the same time. In this work, we propose a novel "CartoonRenderer" framework which utilizing a single trained model to generate multiple cartoon styles. In a nutshell, our method maps photo into a \emph{feature model} and renders the \emph{feature model} back into image space.  In
particular, cartoonization is achieved by conducting some transforma-
tion manipulation in the feature space with our proposed \emph{Soft-AdaIN}. Extensive experimental results show our method produces higher quality cartoon style images than prior arts, with accurate semantic content preservation. In addition, due to the decoupling of whole generating process into \emph{``Modeling-Coordinating-Rendering"} parts, our method could easily process higher resolution photos, which is intractable for existing methods.

\keywords{ Non-photorealistic rendering  \and Neural Style Transfer \and  Image generation.}
\end{abstract}

\section{Introduction}
Cartoon style is one of the most popular artistic styles in today’s world, especially in online social media. To obtain high quality cartoon images, artists
need to draw every line and paint every color block, which is labor intensive.
Therefore, a well-designed approach for automatically rendering realistic photos
in cartoon style is of a great value. Photo cartoonization is a challenging image
stylization task which aims at transforming realistic photos into cartoon style images while preserving the semantic contents of the photos.  Recently, image
stylized rendering has been widely studied and several inspirational methods
have been proposed based on Convolutional Neural Network (CNN). Gatys \cite{Gatys2016Image} formulates image stylized rendering as an optimization problem that translat-
ing the style of an image while preserving its semantic content. 
\begin{figure}[t]
\centering
\includegraphics[width=12.2cm]{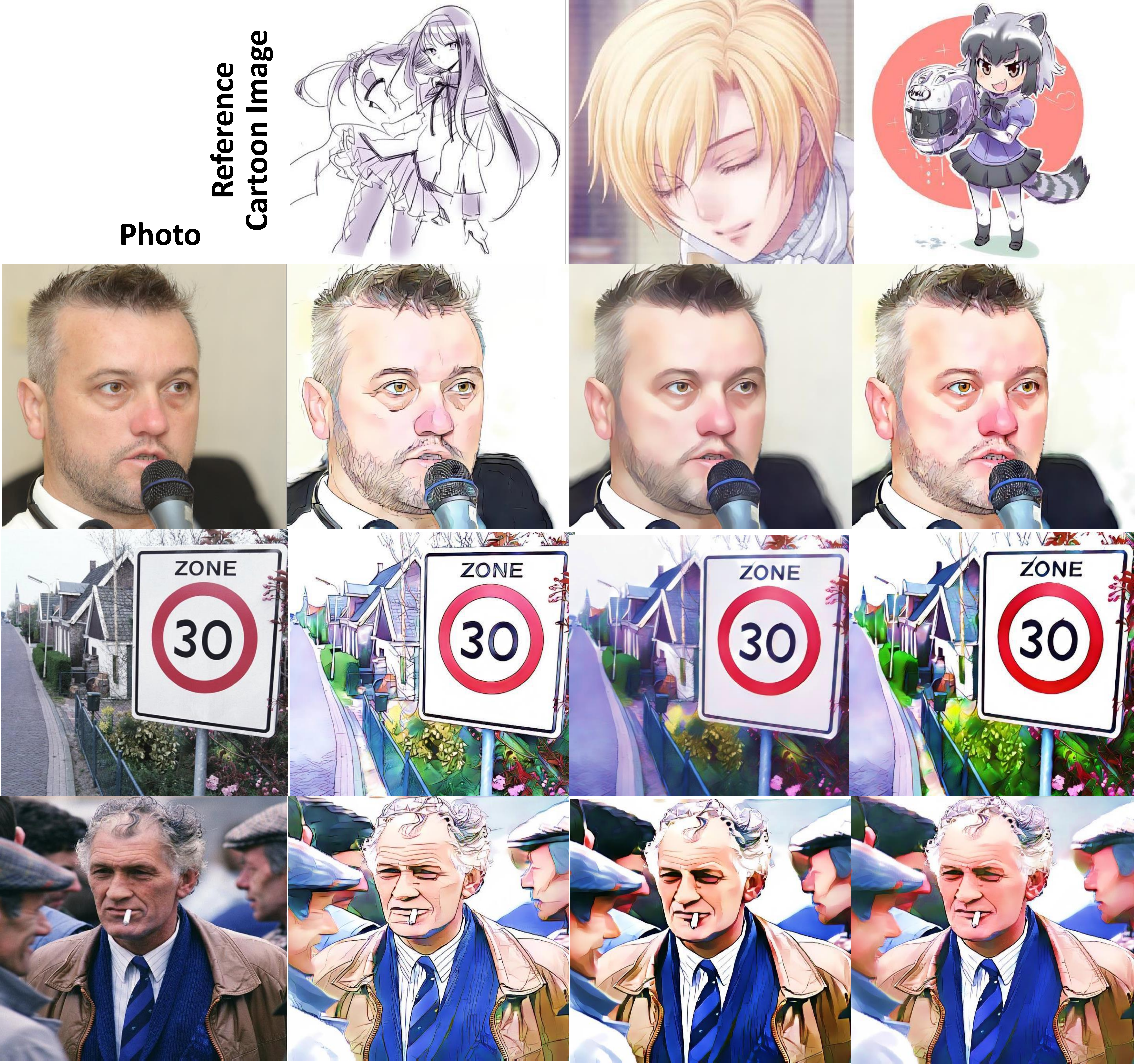}
\caption{Some cartoonization results of \emph{CartoonRenderer} with different reference cartoon images.}
\label{fig1}
\end{figure}
This method produces promising results on transforming images in traditional oil painting styles, such as \emph{Van Gogh}'s style and \emph{Monet}'s style. However, it suffers from long running time caused by tremendous amount of computation. Based on Gatys's pioneering work,  some researchers have devoted substantial efforts to accelerating training and inference process through feed-forward network, such as \cite{Huang2017Arbitrary}\cite{Karras2018A}\cite{Sanakoyeu2018A}\cite{Chen2016Fast}\cite{Yao2019Attention}.  Some methods follow this line of idea that employs a feedforward network as generator to generate stylized results and achieve significant success. Since cartoon style is one of artist styles, many existing methods used for artistic style transfer can also be used to transform realistic photographs into cartoon style. However, even state-of-the-art methods fail to stably produce acceptable results with input content photos in the wild, especially for the high resolution photographs that full of complex texture details. The main reasons are as follows. First, different from artworks in other artistic styles (e.g. oil painting style), cartoon images tend to have clear edges, smooth color blocks and simple textures. As a consequence, cartoon images have sparse gradient information that make it hard for normal convolutional networks to extract valuable features which can well describe cartoon style.  Secondly, clear semantic content is often
difficult to preserve. For example, an apple should still be round and red in car-
toon images. However, current instance-based algorithms tend to preserve local
and noisy details but fail to capture global characteristics of cartoon images.This is because such algorithms purely utilize \emph{Perceptual Loss} or \emph{Gram-Matrix}
to describe image style, and this type of loss encourages to transfer local style
textures which can be described by “strokes” but conflicts to the objective that
preserving detailed semantic contents at the same time. Third, current GAN-
based algorithms cannot handle high resolution images because they utilize an
end-to-end generator which has a large burden in computation .

To address issues mentioned above, we propose \emph{CartoonRenderer}, a novel learning-based approach that renders realistic photographs into cartoon style. Our method takes a set of cartoon images and a set of realistic photographs for training. No correspondence is required between the two sets of training data. It is worth noting that we also do not require cartoon images coming from the same artist.  Similar to other instance based methods, our \emph{CartoonRenderer} receives photograph and cartoon image as inputs and models them respectively in high dimensional feature space to get \emph{feature model"s}. Inspired by AdaIN \cite{Huang2017Arbitrary}, we propose \emph{Soft-AdaIN} for robustly align the \emph{``feature-model"} of photograph according to the \emph{``feature-model"} of cartoon image. Then, we use a \emph{Rendering Network} to generate output cartoonized photograph from the aligned \emph{``feature-model"}. Furthermore, we employ a set of well-designed loss functions to train the \emph{CartoonRenderer}. Except for using pre-trained VGG19 \cite{Simonyan2014Very} model to compute content loss and style loss, we add extra reconstruction loss and adversarial loss to further preserve detailed semantic content of photographs to be cartoonized. Our method is capable of producing high-quality cartoonized photographs, which are substantially better than state-of-the-art methods. Furthermore,  due to the decoupling
of whole generating process into \emph{``Modeling-Coordinating-Rendering"} parts, our method is able to process high resolution photographs (up to 5000*5000 pixels) and maintain high quality of results, which is infeasible for state-of-the-art methods.

\section{Related Works}

\subsubsection{Non-photorealistic rendering (NPR)}
Non-photorealistic rendering is an
alternative to conventional, photorealistic computer graphics, aiming to make
visual communication more effective and automatically create aesthetic results
resembling a variety existing art styles. The main venue of NPR is animation and
rendering. Some methods have been developed to create images with flat shading, mimicking cartoon styles. Such methods use either image filtering or formulations in optimization problem. However, applying filtering or optimization uniformly to the entirely image does not give the
high-level abstraction that an artist would normally do, such as making object
boundaries clear. To improve the results, alternative methods rely on segmentation of images have been proposed, although at the cost of requiring some user
interaction. Dedicated methods have also been developed for portraits, where
semantic segmentation can be derived automatically by detecting facial components. Nevertheless, such methods cannot cope with general images. Turning
photos of various categories into cartoons such as the problem studied in this
paper is much more challenging.

\section{Methodology}
The task of instance based photo cartoonization is to generate cartoon style
version of the given input image, according to some user specified style attributes.
In our method, the style attributes are provided by the reference cartoon image.
Our model learns from a quantity of unpaired realistic photographs and cartoon
images to capture the common characteristics of cartoon styles and re-renders
the photographs into cartoon styles. The photographs and cartoon images are
not required to be paired and the cartoon images are also not required to be
classified by artists. This brings significant feasibility in model training.

For the sake of discussion, let $\mathcal{P}$ and $\mathcal{C}$ be the photograph domain and cartoon domain respectively. Inspired by object modeling and rendering techniques with deep neural networks in the field of computer graphics \cite{Nguyen2018RenderNet}, we formulate photograph cartoonization as a process of \emph{Modeling-Coordinating-Rendering}:

\texttt{Modeling} : $\forall{p} \in \mathcal{P}, \forall{c} \in \mathcal{C}$, we construct the \emph{feature model} of a photograph and a cartoon image as $\Psi_{p}$ and $\Psi_{c}$ respectively.  The \emph{feature model}  consisting of
multi-scale feature maps represent the style characteristics.

\texttt{Coordinating} : Align the \emph{feature model} of photographs $\Psi_{p}$ according to the \emph{feature model} of cartoon image $\Psi_{c}$ and gets the coordinated \emph{feature model} $\Psi_{y}$,  i.e, which possesses $p$’s content representation with $c$’s style representation.

\texttt{Rendering} : Generate the cartoonized photograph $y$ ($y \in \mathcal{C}$) from the coordinated \emph{feature model} $\Psi_{y}$ which can be considered as reconstruction.

In a nutshell, we propose a novel instance based method \emph{``CartoonRenderer"} to render the input photo $p$ into cartoon style result $y$ according to reference cartoon image $c$, which can be described as $y = G(p,c)$. $G$ represents the non-linear mapping function of whole \emph{``Modeling-Coordinating-Rendering"} process. It is worth noting that style attributes are provided by input cartoon image $c$, so single trained \emph{``CartoonRenderer"} can be used to render different cartoon styles by feeding different reference images. Results with different reference cartoon images are demonstrated in Figure \ref{fig1}. We present the detail of our model architecture in Section 3.1 and propose a series of loss functions for training $G$ in Section  3.2.

\begin{figure}[t]
\centering
\includegraphics[width=12.5cm]{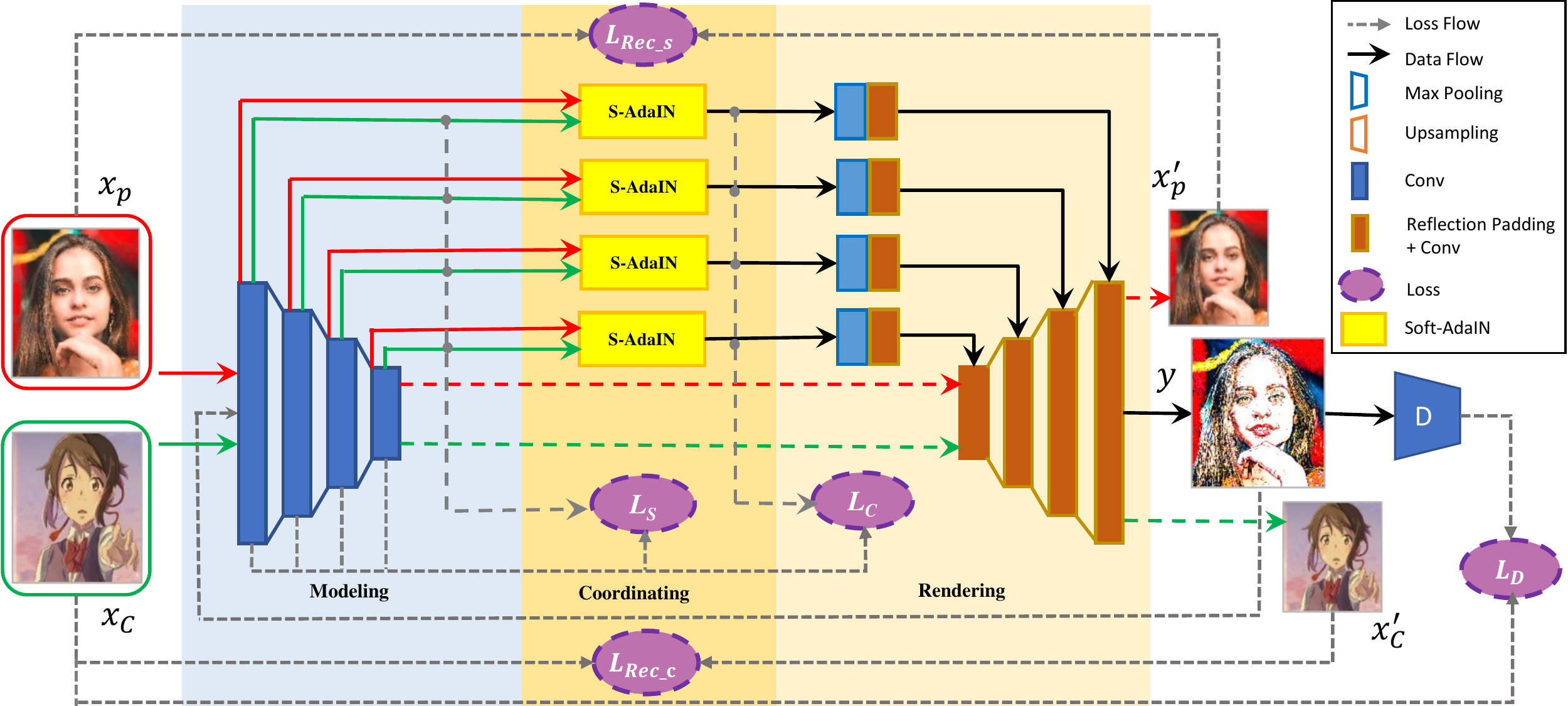}
\caption{The whole framework of our method. $x_p$ represents the input photo and $x_c$ represent the input cartoon image.  $x_p'$ and $x_c'$ represent the reconstructed results and $y$ represents the cartoonized target result.}
\label{fig2}
\end{figure}

\subsection{Model Architecture}
As demonstrated in Figure \ref{fig2}, our \emph{CartoonRenderer} is consist of three parts: Modeling, Coordinating and Rendering. In addition, a discriminator \emph{D} is employed to produce adversarial loss. \emph{CartoonRenderer} follows the auto-encoder architecture. The \emph{Modeling network} is used to map input images into feature spaces. Different from traditional encoder used in Adain \cite{Huang2017Arbitrary} and MUNIT \cite{Huang2018Multimodal}, our \emph{Modeling network} maps input image into multiple scales feature spaces instead of single fixed scale feature space. The \emph{Coordinating} part of \emph{CartoonRenderer} is consist of four \emph{Soft-AdaIN} blocks, corresponding to the number of elements in \emph{feature model}. Each \emph{Soft-AdaIN} block is used to align the corresponding scale’s element in feature models of photo $\Psi_p$ according to the feature model of cartoon image$\Psi_c$.
At last, we train a \emph{Rendering network} to reconstruct back $y$ from the coordinated \emph{feature models} $\Psi_y$.

\subsubsection{Modeling network.}
The \emph{Modeling network} is used to construct the \emph{feature model} of input image. The great success of U-Net \cite{Ronneberger2017U} based methods in high precision segmentation has proved that the coarse contextual information embedded in shallow features plays important role for preserving detailed semantic contents. The shallow features have relatively small receptive field that make it sensitive to local and detailed texture information, meanwhile, the deep features can better at describing global and abstract textures' characteristics. Both local and global texture information are important for generating high-quality images. So we utilize multiple scales of feature responses instead of fixed single scale to represent the images.  The collection of multi-scale features can be
recognized as a high dimensional \emph{feature model}, which contains local and global
semantic information and be able to well-represent the input image in terms of content and style.
We employ the top few layers of a VGG19 \cite{Simonyan2014Very} network (up to $31-th$ layer) as the modeling network. According to the definition of content loss and style loss used in  \cite{Dumoulin2016A}\cite{Ulyanov2016Texture}\cite{Ulyanov2017Improved}, we choose the $4-th$, $11-th$, $18-th$ and $31-th$ layers' output to construct the \emph{feature model}. We define the \emph{feature model} of input image $x$ as:
\begin{align*}
\Psi_x = \{f_4(x),f_{11}(x),f_{18}(x),f_{31}(x)\},
\end{align*}
where $f_{i}(x)$ be the $i-th$ layer's feature response of input $x$. $\Psi_x$ can be recognized as a ``model" of input image $x$ in the space spanned by multi-scale feature subspaces.  The feature model $\Psi_x$ represents image x on different scales.
%------------------------------------------------------------------------------
\subsubsection{\emph{Soft-AdaIN} for robust feature coordinating.}
\emph{AdaIN} proposed in \cite{Huang2017Arbitrary} adaptively computes the affine parameters from the style input instead of learning affine parameters:
\begin{align*}
AdaIN(x_c,x_s) = \sigma(x_s)(\frac{x_c-\mu(x_c)}{\sigma(x_c)})+\mu(x_s).
\end{align*}
\emph{AdaIN} \cite{Huang2017Arbitrary} first scale the normalized content input with $\sigma(x_s)$ and then add $\mu(x_s)$ as a bias. $\sigma(x)$ and $\mu(x)$ are the channel-wise mean and variance of input $x$. As mentioned in Section1, cartoon images tend to have quite unique characteristics which far away from realistic photos, which means that there is a huge gap between the distributions of photos and cartoon images. \emph{AdaIN} \cite{Huang2017Arbitrary} explicitly replace the feature statistics of photo with corresponding feature statistics of cartoon image, which will break the consistency and continuity of feature map. The non-consistency and non-continuity will cause obvious artifacts as shown in Figure \ref{fig3}.c.

To circumvent this problem, we designed \emph{Soft-AdaIN}. The \emph{Soft-AdaIN} is used to align the \emph{feature model} of photographs for cartoon stylizing. As shown in Figure \ref{fig2}, \emph{Coordinating} part consists of four \emph{Soft-AdaIN} blocks. \emph{Soft-AdaIN} also receives a content input and a style input. In this paper, we denote content input (which comes from photographs) as $x_p$ and style input (which comes from cartoon images) as $x_c$. Two mini convolutional networks $\theta_p$ and $\theta_c$ are used to further extract features of $x_p$ and $x_c$. Since the shape of $x_p$ and $x_c$ maybe mismatch, we adopt global average pooling to pool $\theta_p(x_p)$ and $\theta_c(x_c)$ into $1*1*ch$ tensors, where $ch$ is the channel number of $x_p$ and $x_c$. The pooled $\theta_p(x_p)$ and $\theta_c(x_c)$ are concatenated in channel dimension as a $1*1*(2*ch)$ tensor. Then we employ 2 fully-connection layers to compute channel-wise weight $\omega(x_p,x_c)$ from the concatenated tensor. $\omega(x_p,x_c)$ is a $1*1*ch$ tensor. In fact, $\omega(x_p,x_c)$ can be recognized as channel-wise weights for adaptively blend feature statistics of photo input and cartoon input:
\begin{align*}
\sigma'(x_p,x_c) &= \sigma(x_c)*\omega(x_p,x_c) + \sigma(x_p)*(1.0-\omega(x_p,x_c)), \\
\mu'(x_p,x_c) &= \mu(x_c)*\omega(x_p,x_c) + \mu(x_p)*(1.0-\omega(x_p,x_c)).
\end{align*}
Our proposed $Soft-AdaIN$ scales the normalized content input with $\sigma'(x_p,x_c)$ and shift it with $\mu'(x_p,x_c)$:
\begin{align*}
\emph{Soft-AdaIN}(x_p,x_c) = \sigma'(x_p,x_c)(\frac{x_p-\mu(x_p)}{\sigma(x_p)})+\mu'(x_p,x_c).
\end{align*}
We employ \emph{Soft-AdaIN} to coordinate the \emph{feature model} of realistic photo according to \emph{feature model} of cartoon image. For $\forall{p} \in \mathcal{P}, \forall{c} \in \mathcal{C}$, we have
\begin{align*}
\Psi_{p} &= \{f_4(p),f_{11}(p),f_{18}(p),f_{31}(p)\}, \\
\Psi_{c} &= \{f_4(c),f_{11}(c),f_{18}(c),f_{31}(c)\}.
\end{align*}
To perform feature coordination, we conduct $\emph{Soft-AdaIN}$ on each element in $\Psi_{p}$ and $\Psi_{c}$:
\begin{align*}
t^\lambda &= {\emph{Soft-AdaIN}(f_{\lambda}(p),f_{\lambda}(c)),} \\
\Psi_{y} &= {\{t^\lambda\}, \lambda=4,11,18,31,}
\end{align*}
where $\Psi_{y}$ is the coordinated \emph{feature model} for generating output cartoonized result $y$. 

\subsubsection{Rendering Network.}
The \emph{Rendering Network} is used to render \emph{feature model} $\Psi_{x}$ into image space. Our \emph{Rendering Network} has similar architecture as the expansive path of \emph{U-Net} \cite{Ronneberger2017U}. As shown in Figure \ref{fig2}, \emph{Rendering Network} is consist of 4 blocks. The first 3 blocks have two paths: a concatenation path and upsampling path, and the last block only has a upsampling path. The concatenation path receives corresponding scale's element in \emph{feature-model} ($t^{18}$,$t^{11}$ and $t^{4}$ for the first, second and third blocks' concatenation path respectively). The upsampling path receives proceeding block's output (the first block receives $t^{31}$ as input) and is used to expansion the feature responses. We use \emph{Reflection Padding} before each $3*3$ convolutional layer to avoid border artifacts. Obviously, the output of concatenation path and output of upsampling path have the same size, so we can concatenate them along the channel dimension. The concatenated feature map then be fed into another activation layer and become the final output of current block. The last block is a pure upsampling block without concatenation path, and its upsampling path have the same structure as preceding blocks do. All activation functions in \emph{Rendering Network} are \emph{ReLU}. By adopting such multi-scale architecture, the \emph{Rendering Network} is able to make full use of both local and global texture information for generating the output images.

\subsection{Loss Function}
The loss function used to train the \emph{CartoonRenderer} consists of three parts: (1) the style loss $\mathcal{L}_{S}(G)$ which guides the output to have the same cartoon style as the input cartoon image; (2) the content loss $\mathcal{L}_{C}(G)$ which preserves the photographs' semantic content during cartoon stylization; (3) the adversarial loss $\mathcal{L}_{adv}(G,D)$ which further drives the generator network to render input photographs into desired cartoon styles; and (4) the reconstruction loss $\mathcal{L}_{Recon}(G)$ which guides the rendering network to reconstruct origin input images from corresponding un-coordinated \emph{feature models}. We formulate the loss function into a simple additive from:
\begin{align*}
 \mathcal{L}_{adv}(R,D) = \omega_{style}\mathcal{L}_{S}(R) + \omega_{con}\mathcal{L}_{C}(R) + \omega_{recon}\mathcal{L}_{recon}(R) + \omega_{adv}\mathcal{L}_{adv}(R,D),
\end{align*}
where $\omega_{style}$,$\omega_{con}$,$\omega_{recon}$ and $\omega_{adv}$ balance the four losses. In all our experiments. We set $\omega_{recon}=0.0001$, $\omega_{adv}=1$, $\omega_{style}=20$ and $\omega_{con}=1$ to achieve a good balance of style and content preservation.

\subsubsection{Content loss  and style loss.}
 Similar to AdaIN \cite{Huang2017Arbitrary}, we reuse the elements in the \emph{feature model} (feature maps extracted by \emph{modeling network}) to compute the style loss function and content loss function. The \emph{modeling network} is initialized with a pre-trained VGG-19 \cite{Simonyan2014Very}. The content loss is defined as:
 \begin{align*}
 \small {\mathcal{L}_{con}(G) = \sum_{\lambda}\parallel Norm(f_\lambda(G(x,y))) - Norm(f_\lambda(x)) \parallel_1,\lambda=4,11,18,31,}
\end{align*}
 where $Norm(x) = [x-mean(x)]/{var(x)}$ is used to normalize feature maps with channel-wise mean and variance. \\
 Unlike other style transferring methods \cite{Huang2017Arbitrary}\cite{inproceedings}, we define the semantic content loss using the sum of $\mathscr{L}_1$ sparse regularization of normalized VGG feature maps instead of origin feature maps.
 This is due to the fact that statistics of feature maps affect the image's style. Directly using feature maps to compute content loss will introduce some restriction on style, which drives output image very similar to input photograph. The normalization operation eliminates representation of image style from feature maps, so the $\mathscr{L}_1$ sparse regularization of normalized feature maps better describes the differences between semantic contents of photograph and cartoon images. To preserve both local and global semantic contents, we compute $\mathscr{L}_1$ sparse regularization of every feature map in \emph{feature model} and use the sum of them as final content loss. 

 For style loss $\mathcal{L}_{style}(G)$, we adopt the same method as used in AdaIN \cite{Huang2017Arbitrary}. The style loss is defined as:
\begin{align*}
 \mathcal{L}_{style}(G) &= \sum_{\lambda}\parallel \sigma(f_\lambda(G(x,y))) - \sigma(f_\lambda(x))\parallel_2 \\
 &+ \sum_{\lambda}\parallel \mu(f_\lambda(G(x,y))) - \mu(f_\lambda(x))\parallel_2
\end{align*}

\subsubsection{Adversarial loss $\mathcal{L}_{adv}.$}
Because the content loss and style loss mentioned before are both regularization in feature spaces. If there is not explicit restriction in image space, the generated images tend to be inconsistent among different parts and usually contain some small line segments. So we introduce the adversarial training strategy. We use the multi-scale discriminator proposed in \cite{Huang2018Multimodal} to distinguish between real cartoon images and generated cartoonized photographs. $\mathcal{L}_{adv}$ is defined as:
\begin{align*}
 \mathcal{L}_{adv}(G,D) &= E[logD(x_i)] + E[log(1-D(G(x_i,y_j)))].
\end{align*}
The adversarial loss explicitly add restriction to the generated images in image space, which drives the whole generated images more smooth and consistent among different parts. Some ablation study in Section 4.3 proves that the adversarial loss plays important role for producing high-quality cartoonized images.

\subsubsection{Reconstruction loss $\mathcal{L}_{Recon}$.}
The reconstruction loss is consist of two parts: the reconstruction loss for photo images $\mathcal{L}_{Recon\_p}$ and the reconstruction loss for cartoon images $\mathcal{L}_{Recon\_c}$. For input images $x_p$ and $x_c$, we directly render the un-coordinated \emph{feature model} $\Psi_{p}$ and $\Psi_{c}$ and get reconstructed images $x_p'$ and $x_c'$. Reconstruction loss is defined as:
\begin{align*}
 \mathcal{L}_{Recon}(G) &= \mathcal{L}_{Recon\_p}(G) + \mathcal{L}_{Recon\_c}(G), \\
                        &= \parallel x_p' - x_p \parallel_2 + \parallel x_c' - x_c \parallel_2.
\end{align*}
The reconstruction loss is used to ensure the \emph{Rendering network}'s generalization ability. \emph{Rendering network} do not participate in cartoonization process and all cartoonization processes are limited in \emph{Coordinating} part. By adopting reconstruction loss, we make \emph{Rendering network} focus on reconstructing images from \emph{feature models}, which prompts the \emph{Rendering network} be able to render any type image, no matter it is photo ir cartoon image. 

\begin{figure}[t]
\centering
\includegraphics[width=12.2cm]{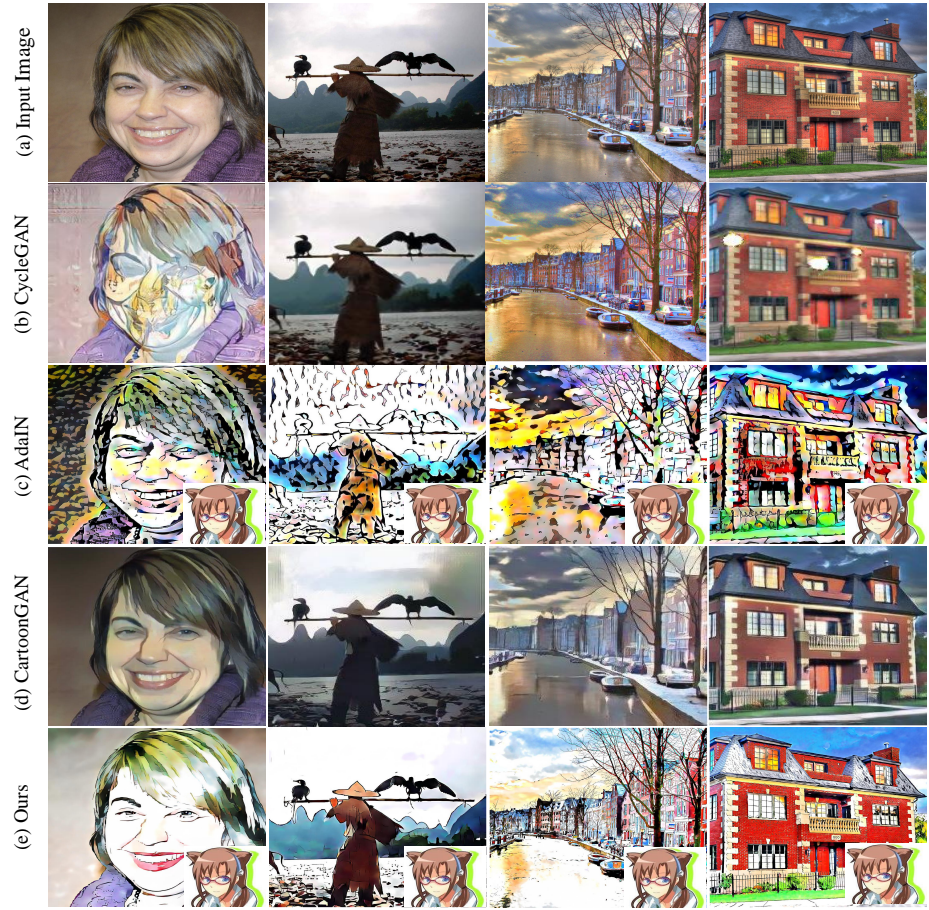}
\caption{Cartoonization results of different methods}
\label{fig3}
\end{figure}
\section{Experiments}
We implement our approach in PyTorch1.0 and Python language. All experiments were performed on an NVIDIA Titan X GPU.

\emph{CartoonRenderer} can generate high-quality cartoonized images by using various cartoon images for training, which are easy to obtain since we do not require paired or classified images. Our model is able to efficiently learn different cartoon sub-styles. Some of results are shown in Figure \ref{fig1}.

To compare \emph{CartoonRenderer} with state of the art methods, we collected
the training and test data as presented in Section 4.1.
In Section 4.2, we present the comparison between \emph{CartoonRenderer} and representative stylization methods. 
%In Section 4.3, we present further ablation experiments to analyze the effectiveness of each component in \emph{CartoonRenderer}.

\subsection{Data}
The training and test data contains realistic photos and cartoon images. All the training images are randomly cropped to 256x256.

\textit{Realistic photos.} We collect 220,000 photos in all, some of which come from MSCOCO \cite{Lin2014Microsoft} dataaset and others come from the Internet. 200,000 photos are used for training and 20,000 for testing. The shortest width of each image is greater than 256 to ensure random cropping is feasible.

\textit{Cartoon images.} We collect 80,000 high-quality cartoon images from the Internet for training, another 10000 cartoon images sampled from Danbooru2018 \cite{danbooru2018} dataset are used for testing.

\subsection{Comparison with state of the art}
In this subsection, we compare our results with CycleGAN \cite{Zhu2017Unpaired}, AdaIN \cite{Huang2017Arbitrary} and CartoonGAN \cite{inproceedings}. 200 epochs are trained for all the them.

Refer to Figure \ref{fig3}, we show the qualitative results generated by different methods, and all of the test data are \textit{never} observed during the training phase. It is clear that CycleGAN \cite{Zhu2017Unpaired} and AdaIN \cite{Huang2017Arbitrary} can not work well with the cartoon styles. In contrast, CartoonGAN and our \emph{CartoonRenderer} produce high-quality results. To preserve the content of the input images well, we add the identity loss to CycleGAN \cite{Zhu2017Unpaired}, but the stylization results are still far from satisfactory.
AdaIN \cite{Huang2017Arbitrary} successfully generates images with smooth colors but suffers from serious artifacts. CartoonGAN \cite{inproceedings} produces clear images without artifacts, but the generated results are too close to the input photos and the color distribution is very monotonous. In other words, the extent of cartoonization with CartoonGAN \cite{inproceedings} is not enough. In contrast, our method apparently produces higher-quality cartoonized images, which have high contrast between colors and contains very clear edges.

\begin{figure}[htbp]
\centering
\subfigure[AdaIN]{
\begin{minipage}[t]{0.25\linewidth}
\centering
\includegraphics[width=1in]{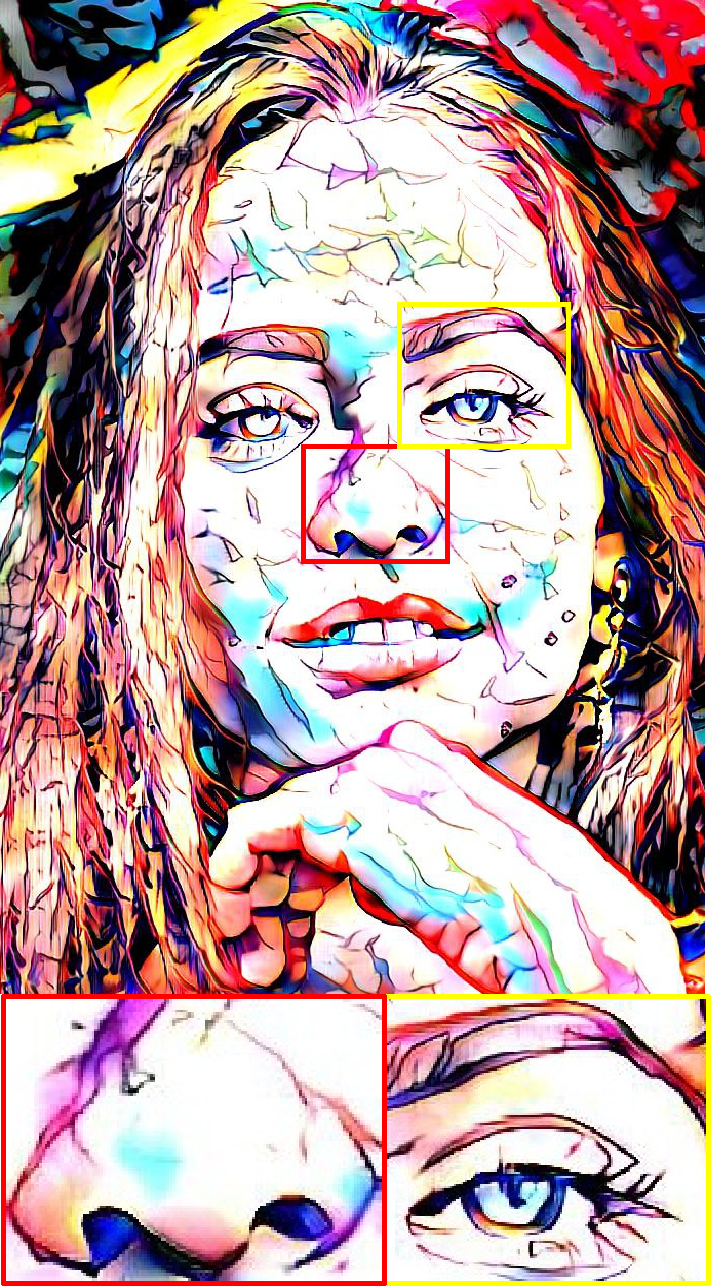}
%\caption{fig1}
\end{minipage}%
}%
\subfigure[CycleGAN]{
\begin{minipage}[t]{0.25\linewidth}
\centering
\includegraphics[width=1in]{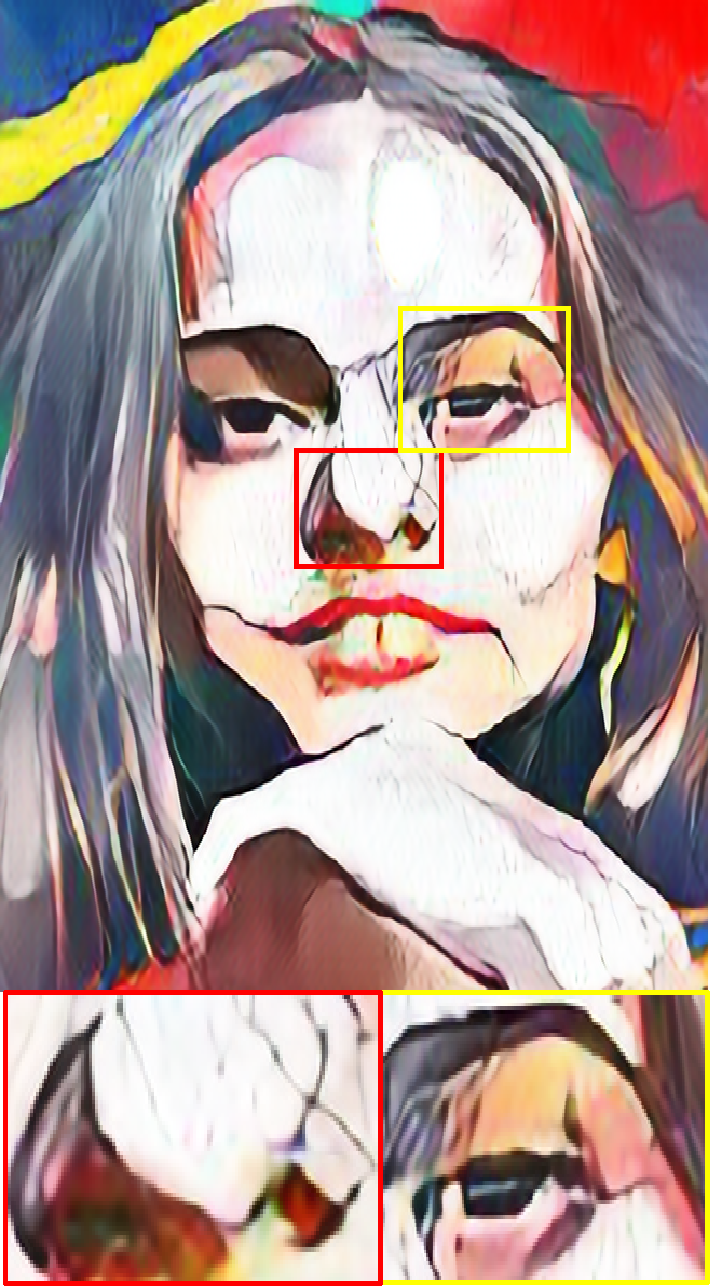}
%\caption{fig2}
\end{minipage}%
}%
\subfigure[CartoonGAN]{
\begin{minipage}[t]{0.25\linewidth}
\centering
\includegraphics[width=1in]{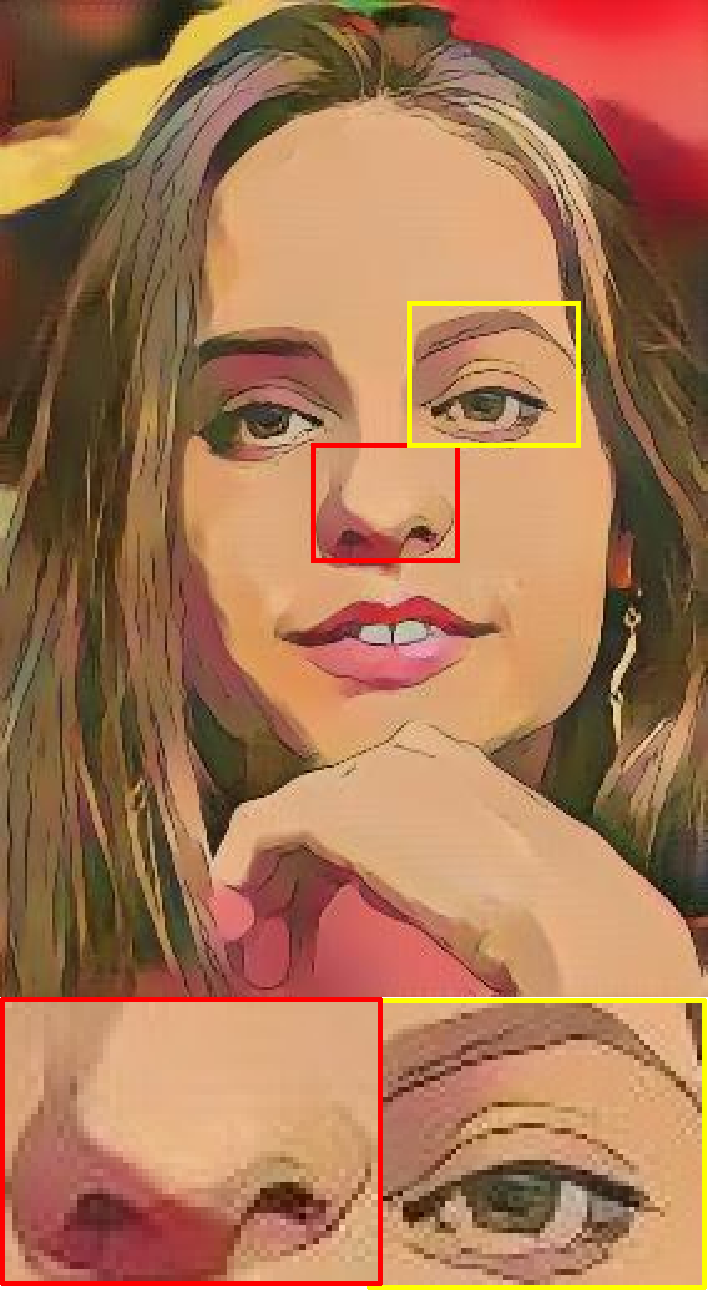}
%\caption{fig2}
\end{minipage}
}%
\subfigure[Ours]{
\begin{minipage}[t]{0.25\linewidth}
\centering
\includegraphics[width=1in]{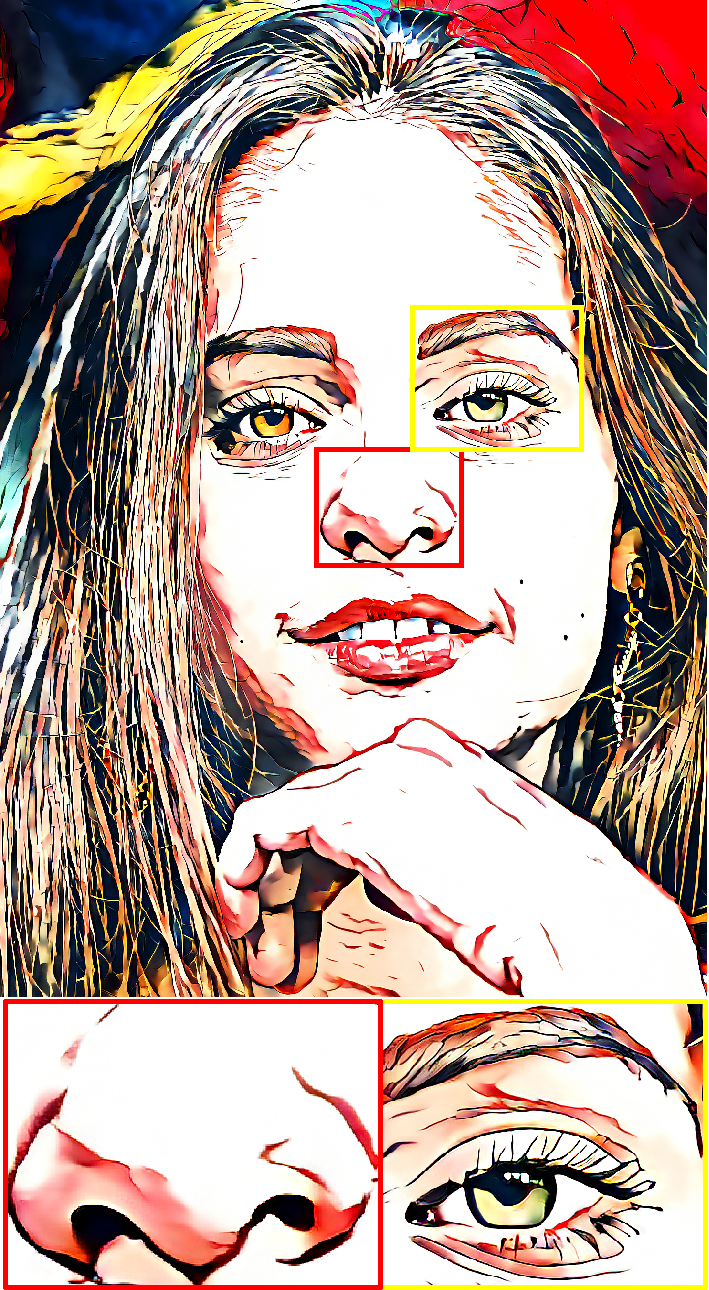}
%\caption{fig2}
\end{minipage}
}%
\centering
\caption{ The detailed performance of different methods}
\label{fig4}
\end{figure}

For more details, we show close-up views of one result in Figure \ref{fig4}. Obviously, our method performs much better than others in detail. Even the details of eyelashes and pupils are well preserved and re-rendered in cartoon style.

%\subsection{Ablation studies}

%

\section{Conclusion}
In this work, we propose a novel "CartoonRenderer" framework which utilizing a single trained model to generate multiple cartoon styles. In a nutshell, our method maps photo into a \emph{feature model} and render the \emph{feature model} back into image space. Our method is able to produce higher quality cartoon style images than prior art. In addition, due to the decoupling of whole generating process into \emph{``Modeling-Coordinating-Rendering"} parts, our method could easily process higher resolution photos (up to 5000x5000 pixels).

\section*{Acknowledgment}
This work was supported by National Natural Science Foundation of China (61976137, U1611461),  111 Project (B07022 and Sheitc No.150633) and the Shanghai Key Laboratory of Digital Media Processing and Transmissions.
This work was also supported by SJTU-BIGO Joint Research Fund, and CCF-Tencent Open Fund.

\newpage
%\begin{figure}
%\includegraphics[width=\textwidth]{fig1.eps}
%\caption{A figure caption is always placed below the illustration.
%Please note that short captions are centered, while long ones are
%justified by the macro package automatically.} \label{fig1}
%\end{figure}

%\begin{theorem}
%This is a sample theorem. The run-in heading is set in bold, while
%the following text appears in italics. Definitions, lemmas,
%ropositions, and corollaries are styled the same way.
%\end{theorem}
%
% the environments 'definition', 'lemma', 'proposition', 'corollary',
% 'remark', and 'example' are defined in the LLNCS documentclass as well.
%
%\begin{proof}
%Proofs, examples, and remarks have the initial word in italics,
%hile the following text appears in normal font.
%\end{proof}
%For citations of references, we prefer the use of square brackets
%and consecutive numbers. Citations using labels or the author/year
%onvention are also acceptable. The following bibliography provides
% sample reference list with entries for journal
%rticles~\cite{ref_article1}, an LNCS chapter~\cite{ref_lncs1}, a
%book~\cite{ref_book1}, proceedings without editors~\cite{ref_proc1},
%and a homepage~\cite{ref_url1}. Multiple citations are grouped
%\cite{ref_article1,ref_lncs1,ref_book1},
%\cite{ref_article1,ref_book1,ref_proc1,ref_url1}.
%
% ---- Bibliography ----
%
% BibTeX users should specify bibliography style 'splncs04'.
% References will then be sorted and formatted in the correct style.
%
\bibliographystyle{splncs04}
\bibliography{samplepaper}
\end{document}